\documentclass[runningheads]{llncs}
\usepackage{graphicx}
\usepackage{subcaption,siunitx,booktabs}

\usepackage{amsmath}
\usepackage{multirow}
\usepackage{siunitx}
\usepackage{booktabs}
\usepackage{subcaption}
\usepackage[maxfloats=256]{morefloats}
\usepackage{algorithm, algorithmicx}
\usepackage{xcolor}
\usepackage{pgf}
\usepackage{algorithm, algpseudocode}

\begin{document}
	\title{An analysis of deep neural networks for predicting trends in time series data\thanks{Supported by the Centre for Artificial Intelligence Research.}}
	%
	%
	\author{Kouame Kouassi, 
		Deshendran Moodley 
	}
	\authorrunning{K. Kouassi and D. Moodley}
	%
	\institute{
		\email{ksskou001@myuct.ac.za,}
		\email{deshen@cs.uct.ac.za}
		\\University of Cape Town and Centre for Artificial Intelligence Research
	}
	\maketitle              
	
\begin{abstract}

     Recently, a hybrid Deep Neural Network (DNN) algorithm, TreNet was proposed for predicting trends in time series data.  While TreNet was shown to have superior performance for trend prediction to other DNN and traditional ML approaches, the validation method used did not take into account the sequential nature of time series data sets and did not deal with model update. In this research we replicated the TreNet experiments on the same data sets using a walk-forward validation method and tested our optimal model over multiple independent runs to evaluate model stability. We compared the performance of the hybrid TreNet algorithm, on four data sets to vanilla DNN algorithms that take in point data, and also to traditional ML algorithms. We found that in general TreNet still performs better than the vanilla DNN models, but not on all data sets as reported in the original TreNet study. This study highlights the importance of using an appropriate validation method and evaluating model stability for evaluating and developing machine learning models for trend prediction in time series data. 

\end{abstract}
	
	\section{Introduction}
With the advent of low cost sensors and digital transformation, time series data is being generated at an unprecedented speed and volume in a wide range of applications in almost every domain.
For example, daily fluctuations of the stock market, traces
produced by a computer cluster, medical and biological experimental
observations, readings obtained from sensor networks, position
updates of moving objects in location-based services, etc, are all
represented in time series. Consequently, there is an enormous interest
in analyzing (including query processing and mining) time
series data, which has resulted in a large number of works on new
methodologies for indexing, classifying, clustering, summarizing, and predicting time series data.

In certain time series prediction applications, segmenting the time series into a sequence of trends and predicting the slope and duration of the next trend is preferred over predicting just the next value in the series \cite{Wang2011,Lin2017}. Piecewise linear representation \cite{Keogh2001a} or trend lines can provide a better representation for the underlying semantics and dynamics of the generating process of a non-stationary and dynamic time series \cite{Wang2011,Lin2017}. Moreover, trend lines are a more natural representation for predicting change points in the data, which may be more interesting to decision makings. For example, suppose a share price in the stock market is currently rising. A trader in the stock market would ask “How long will it take and at what price will the share price peak and when will the price start dropping.” Another example application is for predicting daily household electricity consumption.  Here the user may be more interested in identifying the time, scale and duration of peak energy consumption.

While deep neural networks (DNNs) has been widely applied to computer vision, natural language processing (NLP) and speech recognition, there is limited research on applying DNNs for time series prediction.
In 2017, Lin et al. \cite{Lin2017} proposed a novel approach to directly predict the next trend of a time series as a piecewise linear approximation (\textit{trend line}) with a slope and a duration using a hybrid neural networks dubbed TreNet. The authors of TreNet reported that it outperforms SVR, CNN, LSTM, pHHM \cite{Wang2011}, and cascaded CNN and RNN. However, their study made use of standard cross-validation with random shuffling and a single hold-out set. The use of cross-validation with shuffling implies that data instances, which are generated after the given validation set, are used for training. Besides, the use of a single hold-out set does not provide a sufficiently robust performance measure for data sets that are erratic and non-stationary \cite{Bergmeir2012}.

Furthermore, DNNs, as a result of random initialisation and possibly other random parameter settings could yield substantially different results when re-run with the same hyperparameter values on the same data set. In real world applications where systems are often dynamic, DNN models become outdated and must be frequently updated as new data becomes available. It is also crucial that optimal DNN configurations should be stable, i.e. have minimal deviation from the mean test loss across multiple runs. There is no evidence that this was done for TreNet.  Furthermore, many important implementation details in the TreNet study are not stated explicitly. For instance, the segmentation method used to transform the raw time series into trend lines is not apparent. 
Finally, ensemble regression models such as random forests (RF), and Gradient Boosting Machines (GBM) - which are very widely used traditional machine learning models \cite{Kumar2018,Sharma2017} - are not included in the baseline algorithms. This paper addresses the highlighted shortcomings. Our research questions therefore are:
\begin{enumerate}
   \item Does the hybrid deep neural networks approach for trend prediction perform better than vanilla deep neural networks?
	\item Do deep neural networks models perform better for trend prediction than simpler traditional machine learning (ML) models?
	\item Does the addition of trend line features improve performance over local raw data features alone?
\end{enumerate}
The remainder of the paper is structured as follows. We first provide a brief background of the problem and a summary of related work, followed by the experimental design and give a brief overview of the experiments. We then describe the experiments, present and discuss their results. Finally, we provide a summary and discussion of the key findings.

\section{Background and related work}
\subsection{Background}
The time series trend prediction problem is concerned with predicting the future evolution of the time series from the current time. This evolution is approximated as a succession of time-ordered piecewise linear approximations. The linear approximations indicate the \textit{direction}, the \textit{strength}, and the \textit{length} of the upward/downward movement of the time series. The \textit{slope} of the linear approximation determines the \textit{direction} and the \textit{strength} of the movement, and the number of time steps covered by that linear approximation, i.e. its \textit{duration} determines its \textit{length}.  The formal problem definition is given below.

\subsubsection{Problem formulation}
We define a univariate time series as $X = \{x_1, ..., x_T\}$, where $x_t$ is a real-valued observation at time $t$. The trend sequence $T$ for $X$, is denoted by $T = \{<l_1, s_1>, ..., <s_k, l_k>\}$, and is obtained by performing a piecewise linear approximation of $X$ \cite{Keogh2001a}. $l_k$ represents the  \textit{duration} and is given by the number of data points covered by trend \(k\) and $s_k$ is the slope of the trend expressed as an angle between -90 and 90 degrees. Given a historical time series $X$ and its corresponding trend sequence $T$, the aim is to predict the \textit{next trend} $<s_{k + 1}, l_{k + 1}>$.
	

\subsection{Related work}
Traditional trend prediction approaches include Hidden Markov Models (HMM)s \cite{Wang2011,Matsubara2014} and multi-step ahead predictions \cite{Chang2012}. Leveraging the success of CNNs, and LSTMs in computer vision and natural language processing \cite{LeCun1998,Chung2014EmpiricalEO,Guo2016}, Lin et al. \cite{Lin2017} proposed a hybrid DNN approach, TreNet, for trend prediction. TreNet uses a CNN which take in recent point data, and a LSTM which takes in historical trend lines to extract local and global features respectively. These features are then fused to predict the next trend. While the authors report a marked performance improvement when compared to other approaches, the validation method used in their experiments is questionable. More specifically it does not take into account the sequential nature of times series data. The data was first randomly shuffled, 10\% of the data was held out for testing and a cross validation approach for training with the remainder of the data. Randomly shuffling the data and using a standard cross validation approach does not take into account the sequential nature of time series data and may give erroneous results \cite{Lin2017}.  A walk-forward validation with overlapping partition \cite{LUO2013} is better suited for evaluating and comparing model performance on time series data \cite{LUO2013}. Thus, we replicated the TreNet approach using a walk forward validation over random shuffling and cross validation in this paper. Some follow-up research to TreNet added attention mechanisms \cite{Zhao2018,Feng2019}, however, they not deal with trend prediction specifically. Another active and related field to trend prediction is the stock market direction movement, which is only concerns with the direction of the time series, it does not predict the strength and the duration of the time series \cite{Feng2019,Wen2019,Guo:2019,Liu2018,Nelson2017,Kara2011}. Generally, the baseline methods used by prior works include neural networks, the naive last value prediction, ARIMA, SVR \cite{Lin2017,Zhang2018}. They do not include ensemble methods such as random forest, which are widely used particularly for stock market movement prediction\cite{Kumar2018,Sharma2017}.
	
\section{Experimental design}
	
	
	\subsection{Datasets}\label{sec:exp-datasets}
    Experiments were conducted on the four different datasets described below.
	\begin{enumerate}
		\item The \textit{voltage dataset} from the UCI machine learning repository\footnotemark{}. It contains 2075259 data points of a household voltage measurements of one minute interval. It is highly volatile but normally distributed. It follows the same pattern every year, according to the weather seasons as shown in the appendix figure~\ref{fig:data-voltage}. It corresponds to the power consumption dataset used by Lin et al. \cite{Lin2017}. 
		
		\item The \textit{methane dataset} from the UCI machine learning repository\footnotemark{}.  We used a resampled set of size of 41786 at a frequency of 1Hz.
		The methane dataset is skewed to the right of its mean value and exhibits very sharp changes with medium to low volatility as shown in the appendix figure~\ref{fig:data-methane}. It is corresponds to the gas sensor dataset used by Lin et al. \cite{Lin2017}.
		
		\item 	The \textit{NYSE dataset} from Yahoo finance\footnotemark{}. It contains 13563 data points of the composite New York Stock Exchange (NYSE) closing price from 31-12-1965 to 15-11-2019. 
		Its volatility is very low initially until before the year 2000 after which, it becomes very volatile. 	It is skewed to the right as shown in the appendix figure~\ref{fig:data-nyse}.
		It corresponds to the stock market dataset used by Lin et al. \cite{Lin2017}. 
		
		\item The \textit{JSE dataset} from Yahoo finance. It contains 3094 data points of the composite Johannesburg Stock Exchange (JSE) closing price from 2007-09-18 to 2019-12-31. 
		Compared to the NYSE, this stock market dataset is less volatile and shows a symmetrical distribution around its mean value. However,  it has a flat top and heavy tails on both sides as shown in the appendix figure~\ref{fig:data-jse}.
	\end{enumerate}
	The characteristics of the four datasets are summarised in table~\ref{tab:exp-dataset-characteristics}.
	\begin{table}[!htbp]
		\caption{Summary of the characteristics of the datasets.}
		\label{tab:exp-dataset-characteristics}
		\centering
		\resizebox{0.8\textwidth}{!}{
			\begin{tabular}{lSSSS}
				\toprule
				& {\textbf{\textit{Seasonality}}} & {\textbf{\textit{Periodicity}}} & {\textit{\textbf{Skewness}}} & {\textbf{\textit{Volatility}}}\\
				\midrule
				{\textit{Voltage}}  & {seasonal} & {yearly} & {symmetric} & {very high}\\
				\midrule
				{\textit{Methane}} & {non-seasonal} & {N/A} & {right skewness} & {medium to low}\\	
				\midrule 	
				{\textit{NYSE}} & {non-seasonal} & {N/A} & {right skewness} & {low to high}\\
				
				\midrule
				
				{\textit{JSE}} & {non-seasonal} & {N/A} & {almost symmetric} & {medium to low}\\
				
				\bottomrule	
			\end{tabular}
		}
	\end{table}
	\addtocounter{footnote}{-3} 
	\stepcounter{footnote}\footnotetext{https://archive.ics.uci.edu/ml/datasets/individual+household+electric+power+consumption}
	\stepcounter{footnote}\footnotetext{https://archive.ics.uci.edu/ml/datasets/gas+sensor+array+under+dynamic+gas+mixtures}
	\stepcounter{footnote}\footnotetext{https://finance.yahoo.com}
	
	\subsection{Data preprocessing} \label{sec-exp-preproc}
	The data preprocessing consists of three operations: missing data imputation, the data segmentation, and the sliding window operation.
	Each missing data point is replaced with the closest preceding non-missing value.
	The segmentation of the time series into trend lines i.e. piecewise linear approximations is done by regression using the bottom-up approach, similar to the approach used by \cite{Wang2011}. The data instances, i.e. the input-output pairs are formed using a sliding window. The input features are the local data points ${L_k} = <x_{t_k-w}, ...,x_{t_{k}}>$ for the current trend $T_k = <s_k, l_k>$ at the current time $t$. The window size $w$ is determined by the duration of the first trend line.  The output is the next trend $T_{k+1} = <s_{k+1}, l_{k+1}>$. 
	The statistics of the segmented datasets are provided in the appendix table~\ref{tab:exp-dataset-stats}.
	
	\subsection{Learning algorithms}\label{sec-exp-algorithm}
	The configuration of the seven ML algorithms used in the experiments are now described. They consists of the recent hybrid TreNet \cite{Lin2017}; the vanilla DNN algorithms, i.e. multilayer perceptrons (MLP), the long short-term memory recurrent neural networks (LSTM-RNN), the convolutional neural networks (CNN); and the traditional ML algorithms, i.e. the random forest (RF) regressor, the gradient boosting machine (GBM) regressor, and the radial-based support vector regressor (SVR).
	
	\subsubsection{TreNet} \label{sec:trenet-design}
	\begin{figure}[!htbp]
		\centering
		\resizebox{0.80\textwidth}{!}{
			\includegraphics{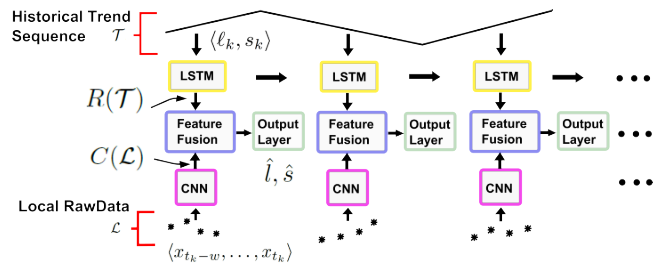}
		}
		\caption{Illustration of the hybrid neural network architecture \cite{Lin2017}}
		\label{trenet}
	\end{figure}
	has a hybrid CNN which takes in raw point data, and LSTM which takes in trend lines as shown in figure~\ref{trenet}. The LSTM consisted of a single LSTM layer, and the CNN is composed of two stacked \cite{Lin2017} 1D convolutional neural networks without pooling layer. The second CNN layer is followed by a ReLU activation function.
	Each of the flattened output of the CNN's ReLU layer and the LSTM layer is projected to the same dimension using a fully connected layer for the fusion operation. The fusion layer consists of a fully connected layer that takes the element-wise addition of the projected outputs of the CNN and LSTM components as its input, and outputs the slope and duration values. A dropout layer is added to the layer before the output layer. The optimal TreNet hyperparameters for each dataset are shown in the appendix table~\ref{tab:trenet-hyper} and compared to Lin et al. \cite{Lin2017}.
	
	\subsubsection{Vanilla DNN algorithms}
	\begin{itemize}
	    \item \textit{The MLP} consists of $N$ number of fully connected neural network (NN) layers, where, $N \in [1, 5]$. Each layer is followed by a ReLU activation function to capture non-linear patterns. To prevent overfitting, a dropout layer is added after each odd number layer, except the last layer. For instance, if the number of layers $N = 5$, the layer 1 and layer 3 will be followed by a dropout layer.
	    
    	\item \textit{The LSTM} consists of $N$ LSTM layers, where $N \in [1, 3]$. Each layer is followed by a ReLU activation function to extract non-linear patterns, and a dropout layer to prevent overfitting. After the last dropout layer, a fully connected NN layer is added. This layer takes the feature representation extracted by the LSTM layers as its input and predicts the next trend. The LSTM layers are not re-initialised after every epoch.
	
	    \item \textit{The CNN} consists of $N$ 1D-convolutional layer, where $N \in [1, 3]$. Each convolutional layer, which consists of a specified number of filters of a given kernel size, is followed by a ReLU activation function, a pooling layer, and a dropout layer to prevent overfitting. The final layer of the CNN algorithm is a fully connected neural network which takes the features extracted by the convolution, activation, pooling, and dropout operations as its input and predicts the next trend. 	The structure of a one layer CNN is illustrated in figure~\ref{fig:cnn-structure}.
    	\begin{figure}[!htbp]
    		\centering
    		\resizebox{0.9\textwidth}{!}{
    			\includegraphics{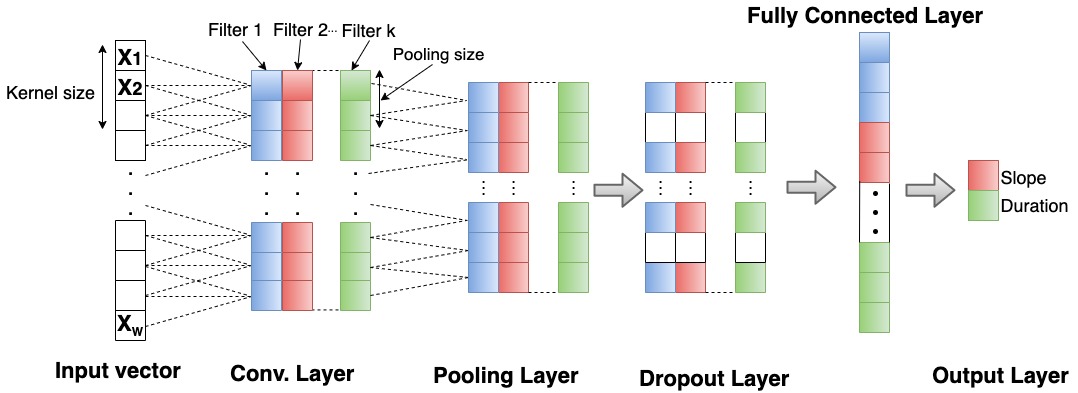}
    		}
    		\caption{The structure of a one layer 1D-convolution neural network.}
    		\label{fig:cnn-structure}
    	\end{figure}
    	Both the convolutional and fully connected layers are initialised with the He initialisation technique \cite{he2015delving}.
\end{itemize}
The vanilla DNN algorithms were also tuned by manual experimentation. The optimal values found for each algorithm on each dataset are shown in the appendix table~\ref{tab:hyperparam-vanilla-dl}.

	\subsubsection{DNN algorithm training, and initialisation}
	The equally weighted average slope and duration mean square error (MSE) is used as a loss function during training using the Adam optimizer \cite{kingma2014adam}. To ensure robustness against random initialisation, the DNNs are initialised using the He initialisation technique \cite{he2015delving} with normal distribution, fan-in mode, and a ReLU activation function.
	
	\subsubsection{Traditional ML algorithms}
	SVR and RF are implemented using Sklearn \cite{scikit-learn}, but, GBM is implemented with LightGBM\footnotemark{}. 
	hyperparameters for the SVR; the \textit{number of estimators}, the \textit{maximum depth}, the \textit{boostrap}, and \textit{warm start} hyperparameters for the RF; as well as the \textit{boostrap type}, the \textit{number of estimators}, and the \textit{learning rate} hyperparameters for the GBM,  on the validation sets. Their optimal hyperparameter configurations per dataset are shown in the appendix table~\ref{tab:hyperparam-trad-ml}. We use the MSE loss for the RF.
	
	\addtocounter{footnote}{-1} 
	\stepcounter{footnote}\footnotetext{https://lightgbm.readthedocs.io/en/latest/index.html}
	
	\subsection{Model evaluation }
	The walk-forward evaluation procedure, with the successive and overlapping training-validation-test partition \cite{LUO2013}, is used to evaluate the performance of the models. The input-output data instances are partitioned into training, validation, and test sets in a successive and overlapping fashion \cite{LUO2013}. For the methane and JSE datasets, the combined test sets make up 10\% of their total data instances as per the original TreNet experiments; and 80\% and 50\% for the voltage and NYSE datasets respectively because of their large sizes. The partition sizes for each dataset are given in the appendix table~\ref{tab:data-partition}. We set the number of partitions to 8 for the voltage, 44 for methane, 5 for NYSE and, 101 for the JSE dataset. This determines the number of model updates performed for each dataset. For example, one initial training and 7 (8-1) model updates are performed for the voltage dataset. For DNN models, the neural networks are initialised using the weights of most recent model, during model update. This makes the training of the network faster without compromising its generalisation ability. More details about this technique which we dubbed model update with warm-start is given in the appendix section~\ref{sec:exp-warm-start}. 
	 The average root mean square error (RMSE) given in equation~\ref{eq:rmse} is used as the evaluation metric. 
	\begin{equation}
	\label{eq:rmse}
	\resizebox{1.0\width}{!}{$	RMSE = \sqrt{\frac{1}{T}\sum_{t = 1}^{T}(y_t - y_t^{'})^2}$}
	\end{equation}
	Each experiment is run 10 times because of the stochastic nature some algorithms, which makes them sensible to random initialisation. The results therefore consists of the mean and the standard deviation of these 10 runs, where applicable.
	
	\section{Overview of experiments}
	We performed four experiments on four datasets. In experiment 1, we implement and evaluate a recent hybrid deep neural neural networks (DNN) trend prediction approach, i.e. TreNet \cite{Lin2017}. TreNet uses a hybrid deep learning structure, that combines both an LSTM and a CNN, and takes in a combination of raw data points and trend lines as its input.  In experiment 2, we compared the TreNet results with the performance of vanilla MLP, CNN and LSTM structures on raw point data to analyse the performance improvement when using a hybrid approach with trend lines. In experiment 3, we evaluate the performance of three traditional ML techniques, i.e. SVR, RF, and GBM on raw point data to analyse the performance difference between DNN and non-DNN approaches. In experiment 4, we supplement the raw data features with trend lines features to evaluate the performance improvement over the raw data features alone for both DNN and non-DNN algorithms.
	
	\section{Experiment 1: Replicating TreNet with walk-forward validation}
    We replicated the TreNet approach using a walk forward validation over random shuffling and cross validation in this paper. 
	In order to compare our results with the original TreNet we use a similar performance measure to Lin et al. \cite{Lin2017}.  We measure the percentage improvement over a naive last value model (LVM). The naive last value model simply "takes the duration and slope of the last trend as the prediction for the next one" \cite{Lin2017}. The use of a relative metric makes comparison easier, since the RMSE is scale-dependent, and the trend lines generated in this study may differ from Lin et al.'s \cite{Lin2017}. Lin et al. \cite{Lin2017} did not provide details of the segmentation method they used in their paper. Furthermore, the naive last value model does not require any hyper-parameter tuning, its predictions are stable and repeatable, i.e. does not differ when the experiment is rerun, and is only dependent on the characteristics of the dataset.
	
	Table~\ref{tab:trenet} shows the performance improvement on RMSE values over the LVM achieved by the TreNet implementation on each dataset. They are compared to the performance of the original TreNet on the three datasets they used in their experiments, i.e. the voltage, methane and NYSE datasets.
	\begin{table}[!htbp]
		\caption{Comparison of the slope (S), duration (D), and average (A) RMSE values achieved by our hybrid neural network's performance and Lin et al.'s results. The percentage improvement (\% improv.) over the naive LVM }
		\label{tab:trenet}
		\centering
		\resizebox{0.95\textwidth}{!}{
			\begin{tabular}{lSSSSSSSSS}
				\toprule
				
				& \multicolumn{3}{c}{Voltage} & \multicolumn{3}{c}{Methane} & \multicolumn{3}{c}{NYSE} \\
				\cmidrule(lr{.75em}){2-4}
				\cmidrule(lr{.75em}){5-7}
				\cmidrule(lr{.75em}){8-10}
				& {S} & {D}  & {A} & {S} & {D} & {A} & {S} & {D} & {A}\\
				
				\midrule
				{Our LVM} & {$17.09$} & {$86.51$} & {$51.80$} & {$28.54$} & {$152.86$} & {$90.70$} & {$127.16$} & {$0.33$} & {$63.75$}\\
				
				\midrule
				{Our TreNet} & {$9.25$} & {$62.37$} & {$35.81$} & {$14.87 $} & {$31.25$} & {$23.06$} & {$86.89$} & {$1.23$} & {$44.06$}\\
				
				\midrule
				\textit{Our \% improv.} & {$\textbf{45.87}$} & {$27.90$} & {$ 30.87$} & {$\textbf{47.90}$} & {$\textbf{79.56}$} & {$ \textbf{74.58}$} & {$\textbf{31.67}$} & {$-272.73$} & {$\textbf{30.89}$}\\
				\midrule
				\midrule
				{Lin et al.'s LVM} & {$21.17$} & {$39.68$} & {$30.43$} & {$10.57$} & {$53.76$} & {$32.17$} & {$8.58$} & {$11.36$} & {$9.97$}\\
				\midrule
				{Lin et al.'s TreNet} & {$12.89$} & {$25.62$} & {$19.26$} & {$9.46$} & {$51.25$}  & {$30.36$} & {$6.58$} & {$8.51$} & {$7.55$}\\
				\midrule
				\textit{Lin et al.'s \% improv.} & {$39.11$} & {$\textbf{35.43}$} & {$\textbf{36.71}$} & {$10.50$} & {$4.69$} & {$ 5.63 $} & {$23.31$} & {$\textbf{25.09}$} & {$ 24.27 $}\\
				\bottomrule	
			\end{tabular}
		}
	\end{table}
	The results of our experiment differ substantially from those reported for the original TreNet. Our TreNet models' percentage improvement over the naive LVM is 13.25 (\textbf{74.58}/5.63) and 1.27 (\textbf{30.89}/24.27) times greater than Lin et al.'s \cite{Lin2017}, on the methane and NYSE datasets respectively; but 1.19 (36.71/\textbf{27.90}) times smaller on the voltage dataset. The naive LVM performs better than our TreNet model on the NYSE for the duration prediction. The -272.73 \% decrease in performance is due to two reasons. On one hand, the model training, i.e. the loss minimisation was biased towards the slope loss at the expense of the duration loss. This is because the slope loss significantly greater compared to the duration loss, but, TreNet's loss function weights both equally. On the other hand, the durations of the trends in the NYSE dataset being very similar - with a standard deviation of 0.81 - makes the last value prediction model a favourably competitive model on the duration prediction. \\
	The greater average improvement on the methane and NYSE is attributed to the use of the walk-forward evaluation procedure. The methane and NYSE datasets undergo various changes in the generating process because of the sudden changes in methane concentrations and the economic cycles for the NYSE. Thus, the use of the walk-forward evaluation ensures that the most recent and useful training set is used for a given validation/test set. However, given a
	validation/test set, cross-validation - used by Lin et al. \cite{Lin2017} - takes all the remaining dataset for training, which contains data instances generated by a different process. Thus, the network learns long-range relationships that are not useful for the current test set. 
    The lower improvement of our TreNet model on the voltage dataset is attributed to our use of a smaller window size for the \textit{local raw data} fed into the CNN. We used 19 compared to their optimal value of 700 on the voltage dataset.
	This shows one of the limitations of our replication of TreNet. For each dataset, we used the length of the first trend line as window size of the \textit{local raw data} feature fed into the CNN, instead of tuning it to select the optimal value. The other weakness concerns the use of a sampled version of the methane dataset instead of the complete methane dataset. 
	
	\section{Experiment 2: Trend prediction with vanilla DNN algorithms}\label{sec:vanilla-dl}
	Given that we are now using a different validation method which yields different performances scores to the original TreNet, we checked whether the TreNet approach still outperforms the vanilla DNN algorithms.	We implemented and tested three vanilla DNN models namely a MLP, LSTM, and CNN using only raw local data features.  
	
	Table~\ref{tab:vanilla-dl} shows the average RMSE values for slope and trend predictions achieved by the vanilla DNNs and TreNet on each dataset across 10 independent runs. The deviation across the 10 runs is also shown to provide an indication of the stability of the model across the runs. We use the average slope and duration RMSE values as an overall comparison metric.  The \% improvement is the improvement of the best vanilla DNN model over TreNet. The best model is chosen based on the overall comparison metric.
	\begin{table}[!htbp]
		\caption{Comparison of the RMSE values achieved by the vanilla DNN models and TreNet.  The \% improvement (\% improv.) is the improvement of the best vanilla DNN model over TreNet}
		\label{tab:vanilla-dl}
		\begin{subtable}{0.95\textwidth}
			\centering
			\resizebox{0.95\textwidth}{!}{
				\begin{tabular}{@{} l*{7}{S} @{}}
					\toprule
					& \multicolumn{3}{c}{Voltage} & \multicolumn{3}{c}{Methane}\\
					\cmidrule(lr{.75em}){2-4} \cmidrule(lr{.75em}){5-7} 
					
					& {Slope} & {Duration} & {Average} & {Slope} & {Duration} & {Average}\\
					
					\midrule			
					\textit{MLP} 
					& {$\textbf{9.04} \pm \textbf{0.06}$} & {$62.82 \pm 0.04$}  & {$35.93 \pm 0.05$} 
					& {$14.57\pm 0.10$} & {$49.79 \pm 4.85$}  & {$32.18 \pm 2.48$} \\
					
					\midrule
					{LSTM}
					& {$10.30 \pm 0.0$} & {${62.87} \pm {0.0}$}  & {${36.59} \pm {0.0}$} 
					& {$\textbf{14.21} \pm \textbf{0.19}$} & {$56.37 \pm 1.77$}  & {$35.29 \pm 0.49$} \\
					
					\midrule
					\textit{CNN} 
					& {$9.24 \pm 0.10$} & {$62.40 \pm 0.13$} & {$35.82 \pm 0.12$} 
					& {${15.07} \pm {0.35}$} & {$54.79 \pm 4.55$}  & {$34.93 \pm 2.45$} \\
					\midrule
					\textit{TreNet}
					& {$9.25 \pm 0.0$} & {$\textbf{62.37}\pm \textbf{0.01}$} & {$\textbf{35.81} \pm \textbf{0.01}$} 
					& {$14.87 \pm 0.40$} & {$\textbf{31.25} \pm \textbf{2.62}$} & {$\textbf{23.06} \pm \textbf{1.51}$} \\
					
					\midrule
					\textit{\% improv.} 
					& {-0.11} & {-0.05} & {-0.03}
					& {\textbf{2.02}} & {-59.33} & {-39.55}\\
					
					\bottomrule	
				\end{tabular}
			}
		\end{subtable}
		
		\begin{subtable}{0.95\textwidth}
			\centering
			\resizebox{0.95\textwidth}{!}{
				\begin{tabular}{@{} l*{7}{S} @{}}
					\toprule
					& \multicolumn{3}{c}{NYSE} & \multicolumn{3}{c}{JSE} \\
					\cmidrule(lr{.75em}){2-4} \cmidrule(lr{.75em}){5-7} 
					
					& {Slope} & {Duration} & {Average} & {Slope} & {Duration} & {Average}\\
					
					\midrule
					\textit{MLP}  
					& {$90.76 \pm 4.43$} & {$33.08 \pm 42.08$}  & {$61.92 \pm 23.26$} 
					& {${19.87} \pm {0.01}$} & {$12.51 \pm 0.09$}  &{${16.19} \pm {0.05}$} \\
					
					\midrule
					{LSTM}
					& {$\textbf{86.56} \pm \textbf{0.01}$}& {$\textbf{0.41} \pm \textbf{0.08}$}  & {$\textbf{43.49} \pm \textbf{0.05}$} 
					& {${19.83 }\pm {0.01}$} & {$12.68 \pm 0.01$} & {${16.25} \pm {0.01}$}\\
					
					\midrule
					\textit{CNN} 
					& {$89.31 \pm 1.38$} & {$12.21 \pm 12.17$} & {$50.76 \pm 6.78$} 
					& {${19.90} \pm {0.06}$} & {$\textbf{12.48} \pm \textbf{0.21}$} & {${16.19 }\pm {0.14}$}\\
					
					\midrule
					\textit{TreNet}
					& {$86.89 \pm 0.14$} & {$1.23 \pm 0.38$} & {$44.06 \pm 0.26$} 
					& {$\textbf{19.65} \pm \textbf{0.05}$} & {${12.49} \pm {0.04}$} & {$\textbf{16.07} \pm \textbf{0.05}$}\\
					
					\midrule
					\textit{\% improv.} 
					& {\textbf{0.38}} & {\textbf{66.67}} & {\textbf{1.29}}
					& {-1.12} & {-0.16} & {-0.75}\\
					
					\bottomrule	
			\end{tabular}}
		\end{subtable}
		
	\end{table}
	\\In general TreNet still performs better than the vanilla DNN models, but does not outperform the vanilla models on all the datasets. 
	The most noticeable case is on the NYSE, where the  LSTM model outperforms the TreNet model on both the slope and duration prediction. 
	This contradicts Lin et al. \cite{Lin2017}'s findings, where TreNet clearly outperforms all other models including LSTM. On average, Lin et al.'s \cite{Lin2017} TreNet model outperformed their LSTM model by 22.48\%; whereas, our TreNet implementation underperformed our LSTM model by 1.31\%. However, Lin et al. \cite{Lin2017}'s LSTM model appears to be trained using trend lines only and not raw point data. This LSTM model uses \textit{local raw data} features. It must also be noted that the validation method used here is substantially different from the one used by Lin et al. \cite{Lin2017}. 
	The large performance difference between TreNet and the vanilla models on the methane dataset is because for this dataset the raw local data features do not provide the global information about the time series since it is non-stationary. This is confirmed by the increase in the performance of the MLP (23.83\%), LSTM (11.02\%) and CNN (24.05\%) after supplementing the raw data features with trend line features (see experiment 4 in section~\ref{sec:exp4}).
			
	\section{Experiment 3: Traditional ML models}\label{sec:trad-ml}
	Given the new validation method, we now compare the performance of DNN trend prediction models to the performance of traditional ML models. We implemented and tested three traditional ML models, i.e.  radial-based SVR, RF, and GBM. To our knowledge, RF and GBM have not been used previously for trend prediction. Lin et al. \cite{Lin2017} compared their approach against multiple SVR kernels that took in both local raw data and trend line features. In this experiment, our models take in only \textit{local raw data} features without trend lines. 
	
	Table~\ref{tab:vanilla-dl-trenet} shows the RMSE values achieved by the traditional ML algorithms and the best DNN models on each dataset. The best DNN model is TreNet on all datasets except on the NYSE, on which LSTM is the best model. The improvement (\%) is the performance improvement of the best traditional ML model over the best DNN model, where, the best model is selected based on the equally weighted average slope and duration RMSE, i.e. average. 
	\begin{table}[!htbp]
		\caption{Comparison of the best DNN models (Best DNN) with the traditional ML algorithms. The \% improvement (\% improv.) is the performance improvement of the best traditional ML model over the best DNN model}
		\label{tab:vanilla-dl-trenet}
		\begin{subtable}{0.95\textwidth}
			\centering
			\resizebox{0.95\textwidth}{!}{
				\begin{tabular}{@{} l*{7}{S} @{}}
					\toprule
					& \multicolumn{3}{c}{Voltage} & \multicolumn{3}{c}{Methane}\\
					\cmidrule(lr{.75em}){2-4} \cmidrule(lr{.75em}){5-7} 
					
					& {Slope} & {Duration} & {Average} 
					& {Slope} & {Duration} & {Average}\\

					\midrule
					{\textit{RF}} 
					& {$9.53 \pm 0.0$} & {${63.11} \pm {0.20}$} & {$36.32 \pm 0.10$} 
					& {$\textbf{10.09} \pm \textbf{0.01}$} & {$\textbf{20.79} \pm \textbf{0.01}$}  & {$\textbf{15.44} \pm \textbf{0.01}$}\\
					
					\midrule
					{\textit{GBM}} 
					& {${10.0} \pm {0.0}$} & {$62.67 \pm 0.0$}  & {$36.34 \pm 0.0$} 
					& {$13.05 \pm 0.0$} & {$75.10 \pm 0.0$}  & {$44.08 \pm 0.0$}\\
					
					\midrule
					{\textit{SVR}} 
					& {${{9.32}} \pm {0.0}$} & {${62.58} \pm {0.0}$} & {${35.95} \pm {0.0}$} 
					& {$14.98 \pm 0.0$} & {${34.39} \pm {0.0}$}  & {${24.69} \pm {0.0}$} \\ 
					
					\midrule
					{\textit{Best DNN}} 
					& {$\textbf{9.25} \pm \textbf{0.0}$} & {$\textbf{62.37}\pm \textbf{0.01}$} & {$\textbf{35.81} \pm \textbf{0.01}$} 
					& {$14.87 \pm 0.40$} & {$31.25 \pm 2.62$} & {$23.06 \pm 1.51$} \\
					
					\midrule
					{\textit{\% improv.}} 
					& {$-0.76$} & {$-0.34$} & {$-0.47$} 
					& {$\textbf{32.15}$} & {$\textbf{33.47}$} & {$\textbf{33.04}$} \\
					\bottomrule	
			\end{tabular}}
		\end{subtable}
		\begin{subtable}{0.95\textwidth}
			\centering
			\resizebox{0.95\textwidth}{!}{
				\begin{tabular}{@{} l*{7}{S} @{}}
					\toprule
					& \multicolumn{3}{c}{NYSE} & \multicolumn{3}{c}{JSE} \\
					\cmidrule(lr{.75em}){2-4} \cmidrule(lr{.75em}){5-7} 
					
					& {Slope} & {Duration} & {Average} 
					& {Slope} & {Duration} & {Average}\\
					
					\midrule
					{\textit{RF}} 
					& {$88.75 \pm 0.17$} & {$\textbf{0.29} \pm \textbf{0.0}$}  & {$44.52 \pm 0.09$} 
					& {${20.21} \pm {0.0}$} & {${12.67} \pm {0.0}$}  & {${16.44} \pm {0.0}$}\\
					
					\midrule
					{\textit{GBM}} 
					& {$86.62 \pm 0.0$}& {$0.42 \pm 0.0$}  & {$43.52 \pm 0.0$} 
					& {$20.08 \pm 0.0$} & {${12.62} \pm {0.0}$}  & {$16.35 \pm 0.0$}\\
					
					\midrule
					{\textit{SVR}} 
					& {$\textbf{86.55} \pm \textbf{0.0}$} & {$0.42 \pm 0.0$}  & {${43.49} \pm {0.0}$} 
					& {${20.01} \pm {0.0}$} & {${12.85} \pm {0.0}$}  & {${16.43} \pm {0.0}$}\\ 
					
					\midrule
					{\textit{Best DNN}}  
					& {${86.56} \pm {0.01}$} & {$\textbf{0.41} \pm \textbf{0.08}$} & {${43.49} \pm {0.05}$}
					& {$\textbf{19.65} \pm \textbf{0.05}$} & {$\textbf{12.49} \pm \textbf{0.04}$} & {$\textbf{16.07} \pm \textbf{0.05}$}\\  
					
					\midrule
					{\textit{\% improv.}}  
					& {$\textbf{0.01}$} & {${2.44}$} & {${0.0}$} 
					& {$-2.19$} & {$-1.04$} & {$-1.74$}\\
					\bottomrule	
			\end{tabular}}
		\end{subtable}
		
	\end{table}

	The best traditional ML algorithm underperformed the best DNN algorithm by 0.47\% and 1.74\% respectively on the (almost) normally distributed datasets such voltage and the JSE datasets. However, the RF model outperformed the best DNN model, i.e. TreNet by 33.04\% on the methane dataset; while the SVR model matched the performance of the best DNN model, i.e. LSTM on the NYSE dataset. TreNet learns long-range dependencies from trend line features with its LSTM component. Although this is useful for stationary and less evolving time series such as the voltage and JSE datasets, it appears that it can be detrimental in the case of dynamic and non-stationary time series such as the methane dataset. This may explain why the traditional ML models, which do not keep long-term memory, performed better this dataset.\\
	The fact that the radial-based SVR performed better than TreNet on the NYSE dataset contradicts Lin et al. \cite{Lin2017}'s results. We attribute this to the use the use of \textit{local raw data} features alone, instead of \textit{local raw data} plus \textit{trend line} features used by Lin et al. \cite{Lin2017}.
	
	

	\section{Experiment 4: Addition of trend line features}\label{sec:exp4}
	In this experiment, we supplement the raw data with trend line features to analyse whether this yields any performance improvement to the DNN and non-DNN models from Experiments 2 and 3. We did retain the hyperparameter values found using the \textit{raw data} features alone for this experiment.
	
	Table~\ref{tab:addition-of-trendlines} shows the average performance improvement (\%) after supplementing the raw data with trend line features. The negative sign indicates a drop in performance. The actual RMSE values are shown in the appendix table~\ref{tab:feature-type-dnn} and table\ref{tab:feature-type-ensemble}.
	\begin{table}[!htbp]
		\caption{Performance improvement after supplementing the raw data with trend line features.}
		\label{tab:addition-of-trendlines}
		\centering
					\resizebox{0.6\textwidth}{!}{
		\begin{tabular}{lSSSSSSS}
			\toprule
			& {MLP} & {LSTM} & {CNN} & {RF} & {GBM} & {SVR} & {Average}\\
			
			\midrule
			{\textit{Voltage}} 
			& {\textbf{0.03}} & {0.0} & {-73.14} & {-0.13} & {\textbf{0.06}} & {-0.36} & {-12.26}\\
			
			\midrule
			{\textit{Methane}} 
			& {\textbf{23.83}} & {\textbf{11.02}} & {\textbf{24.05}} & {-4.47} & {\textbf{42.88}} & {-6.28} & {\textbf{15.17}}\\
			
			\midrule
			{\textit{NYSE}} 
			&{\textbf{6.49}} & {0.0} & {-1.00} & {\textbf{2.36}} & {\textbf{0.23}} & {-0.02} & {\textbf{1.34}}\\
			
			\midrule
			{\textit{JSE}} 
			& {{-4.14}} & {{-1.17}} & {-5.37} & {-7.60} & {\textbf{0.37}} & {-10.96} & {-4.81}\\
			
			\midrule
			\midrule
			{\textit{Average}} 
			& {\textbf{6.55}} & {\textbf{4.93}} & {-13.87} & {-2.46} & {\textbf{10.89}} & {-4.41} & \\
			
			\bottomrule	
		\end{tabular}
	}
	\end{table}	
	\\The addition of trend line features improved the performance of both DNN and non-DNN models 10 times out of 24 cases. In general, it improves the performance of dynamic and non-stationary time series such as the methane and NYSE datasets. This is because local raw data features do not capture the global information about the time series for non-stationary time series. Thus, addition of trend line features brings new information to the models. 
	In 12 out of 24 cases, the addition of trend line features reduced the performance of both DNN and non-DNN models except the GBM models. For these cases, the addition trend line features brings noise or duplicate information, which the models did not deal with successfully. This may be because the optimal hyperparameters for the raw data features alone may not be optimal for the raw data and the trend line features combined. For instance, DNN models are generally able to extract the true signal from noisy or duplicate input features, however, they are sensitive to the hyperparameter values. 
	
	The above results show that the addition of trend line features has the potential to improve the performance of both DNN and non-DNN models on non-stationary time series. This comes at the cost of additional complexities and restrictions. The first complexity is related to the model complexity because the bigger the input feature size, the more complex the model becomes. 
	Secondly, the trend line features require the segmentation of the time series into trends, which brings new challenges and restrictions during inference. 
	For instance, trend prediction applications that require online inference need an online segmentation method such as the one proposed by Keogh et al. \cite{Keogh2001a}.
	It is therefore necessary to evaluate whether the performance gain over raw data features alone justifies these complexities and restrictions.

    \section{Summary of the best performing trend prediction algorithms}
    Table~\ref{tab:overall-summary} provides a summary of the best models and their average performance from all four experiments. 
    \begin{table}[!htbp]
	    \centering
		\caption{Average RMSE values (E) achieved by the hybrid algorithm, i.e. TreNet; and the best non-hybrid algorithm (A) with raw point data features alone (Pt) and with raw point data plus trend line features (Pt + Tr).  The \% change is with respect to the TreNet algorithm.}
		\label{tab:overall-summary}
		\resizebox{0.90\columnwidth}{!}{
			\begin{tabular} {llcccccc}
				\toprule
				& & {\textit{\% Change}} & {\textit{Pt}} & {\textit{Hybrid}}  & {\textit{Pt + T}} & {\textit{\% Change}}\\
				\midrule
				\multirow{2}{*}{\textit{Voltage}} 
                & \textit{A} & {-} & {CNN} & {\textbf{TreNet}}  & {MLP} & {-}\\
				& {\textit{E}} & {-0.03} & {${35.82} \pm {0.12}$} & {$\textbf{35.81} \pm \textbf{0.01}$} & {${35.92} \pm {0.05}$} & {-0.31}\\
				\midrule
				\multirow{2}{*}{\textit{Methane}} 
	            & \textit{A} & {-} & {\textbf{RF}}& {TreNet} & {RF} & {-}\\
				& {\textit{E}} & {\textbf{33.04}} & {$\textbf{15.44} \pm \textbf{0.01}$} & {${23.06} \pm {1.51}$} & {${16.13} \pm {0.01}$} & {30.05} \\
					
				\midrule
				\multirow{2}{*}{\textit{NYSE}} 
	            & \textit{A} & {-} & {SVR} & {TreNet}  & {\textbf{GBM}} & {-}\\
				& {\textit{E}} & {\textbf{1.29}} & {${43.49} \pm {0.0}$} & {${44.06} \pm {0.26}$} & {$\textbf{43.42} \pm \textbf{0.0}$} & {1.45}\\
				
				\midrule
				\multirow{2}{*}{\textit{JSE}} 
	            & \textit{A} & {-} & {MLP} & {\textbf{TreNet}}  & {GBM} & {-}\\
				& {\textit{E}} & {-0.75} & {${16.19} \pm {0.05}$}  & {$\textbf{16.07} \pm \textbf{0.05}$} & {${16.29} \pm {0.0}$} & {-1.37}\\
				\bottomrule
			\end{tabular}
		}
	\end{table}
	The TreNet algorithm outperforms the non-hybrid algorithms on the voltage and JSE datasets, but the performance difference is marginal $<1\%$. Interestingly, the traditional ML algorithms outperformed TreNet and the vanilla DNN algorithms on the methane and NYSE datasets. 
	
	The additional of trend lines to the point data (experiment 4) did not yield any substantial change in the results. It must be noted though that this was an exploratory experiment and that no hyper-parameter optimisation was done to cater for the introduction of a new input feature. It may well be the case that better models could be found of a new hyper-parameter optimisation process was undertaken.
	
	It is clear from these results that TreNet generally performs well on most data sets. However, it is not the clear winner, and there are some data set where traditional models can substantially outperform TreNet. It is also clear that models built with point data alone can generally reach the performance levels of TreNet. 
	
	\section{Discussion and Conclusions}

	In this work, we identify and address some limitations of a recent hybrid CNN and LSTM approach for trend prediction, i.e. TreNet. We used an appropriate validation method, i.e. walk-forward validation instead of the standard cross-validation and also tested model stability. We compared TreNet to vanilla deep neural networks (DNNs) that take in point data features. Our results show that TreNet does not always outperform vanilla DNN models and when it does the outperformance is marginal. Furthermore, our results show that for non-normally distributed datasets, traditional ML algorithms, such as Random forests and Support Vector Regressors, can outperform more complex DNN algorithms. We highlighted the importance of using an appropriate validation strategy and testing the stability of DNN models when they are updated and retrained as new observations becomes available.
	
	There are many avenues to probe the results of this work further. Firstly we only tested this on four data sets. While these included all three data sets used in the original TreNet paper \cite{Lin2017}, testing on more data sets is required to probe the generalisation of these findings. Secondly, there are some avenues that can be explored to improve on these results. Since the window size was fixed to the duration length of the first trend line, the effect of varying the window size could be tested.  A sampled version of the methane dataset is used instead of the complete methane dataset and the full data set could be used. Finding the optimal hyper-parameter values for a particular time series required extensive experimentation, and often requires information about the characteristics of that time series. Automatic machine learning techniques such as \cite{Falkner2018,Li} could be explored to search for automating algorithm selection and hyper-parameter configuration. configuration.

	\bibliographystyle{splncs04}

	\section{Appendices}
	\begin{figure}
		\centering
		\resizebox{0.8\textwidth}{!}{
			\includegraphics[width=\textwidth]{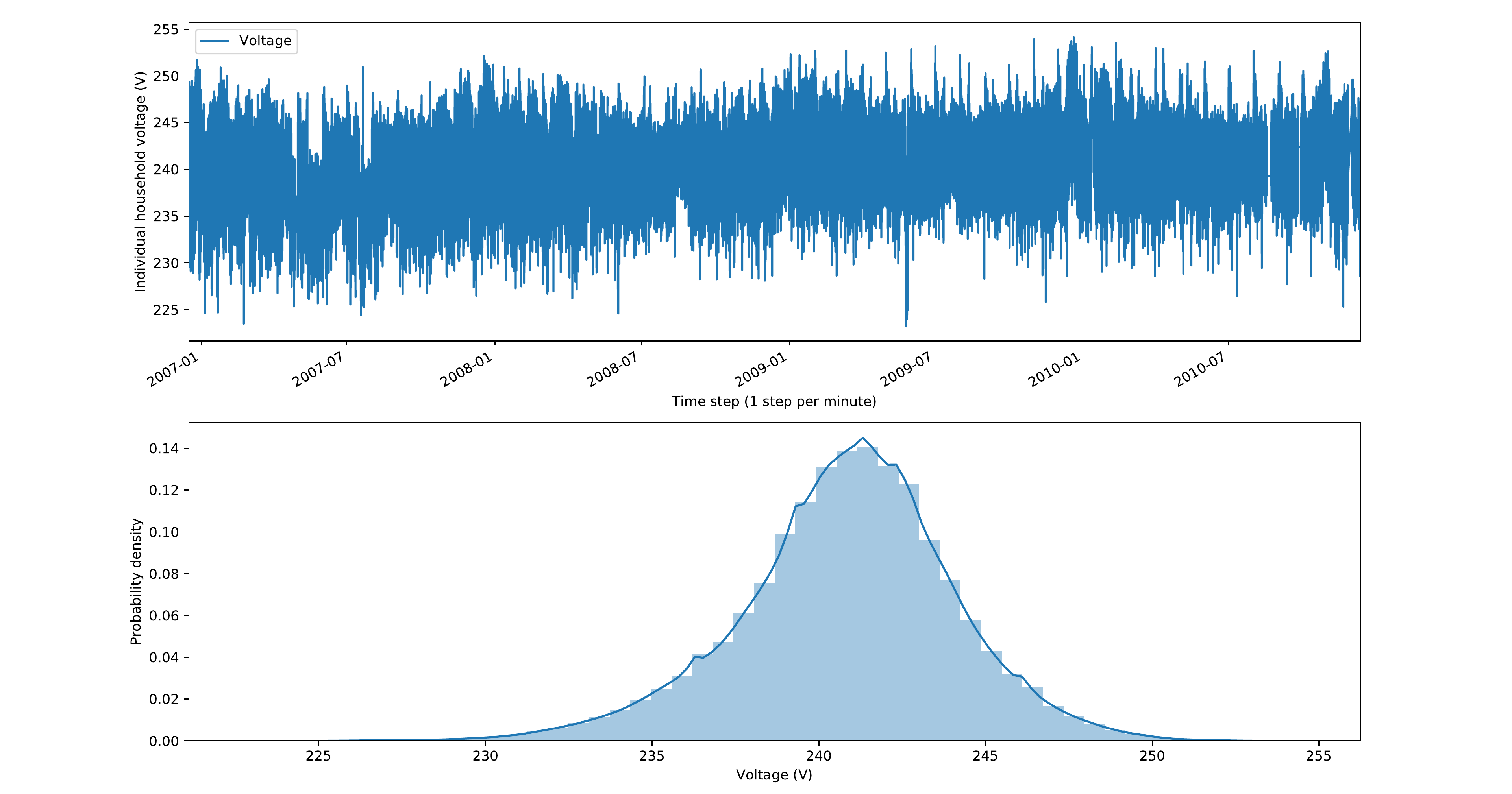}
		}
		\caption{Top - The individual household voltage dataset.  Bottom - Probalility distribution of the voltage dataset.}
		\label{fig:data-voltage}
	\end{figure}
	
	\begin{figure}
		\centering
		\includegraphics[width=\textwidth]{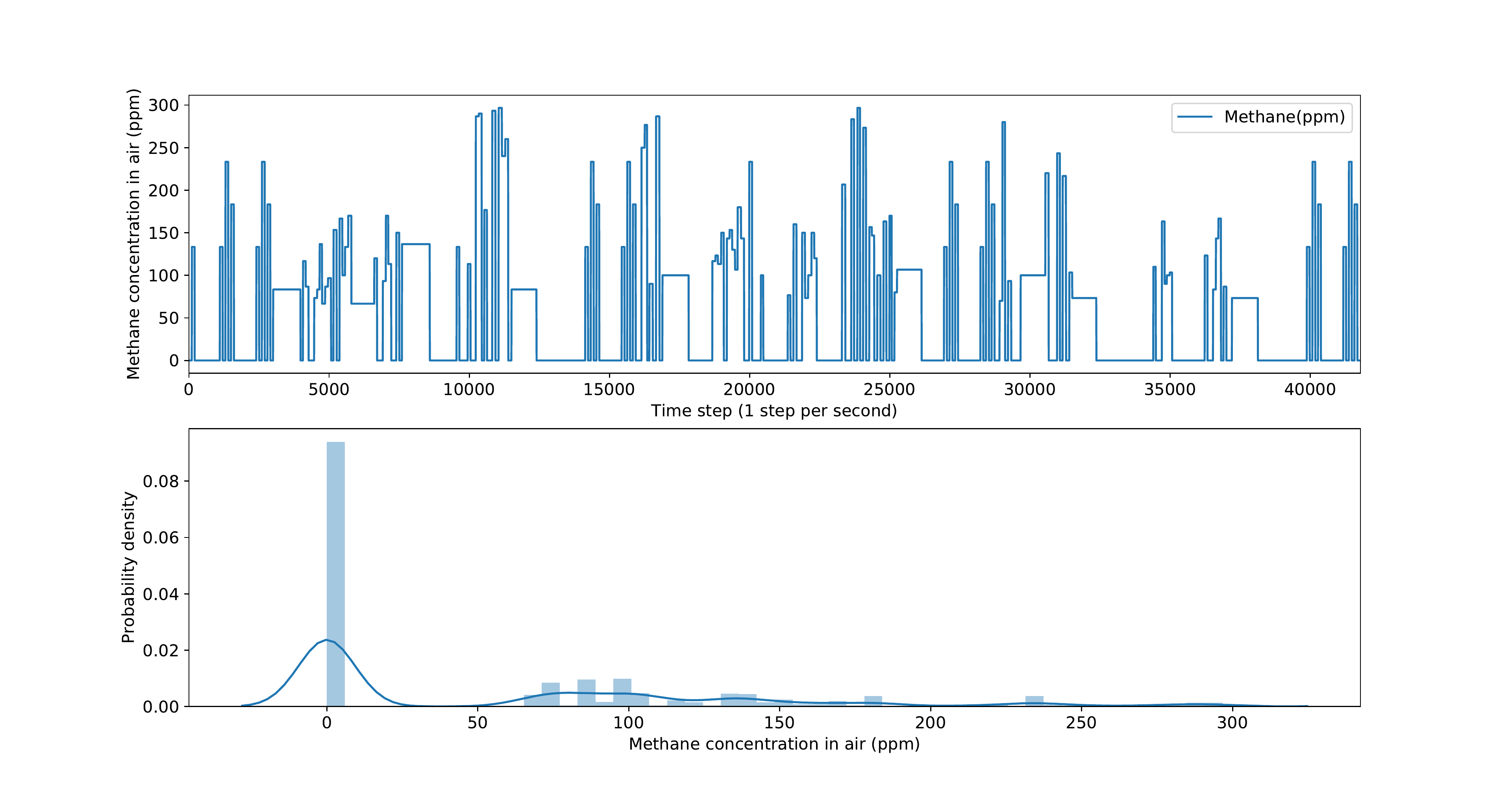}
		\caption{Top - Methane concentration in air over time. Bottom - Probability distribution of the methane dataset.}
		\label{fig:data-methane}
	\end{figure}
	
	\begin{figure}
		\centering
		\includegraphics[width=\textwidth]{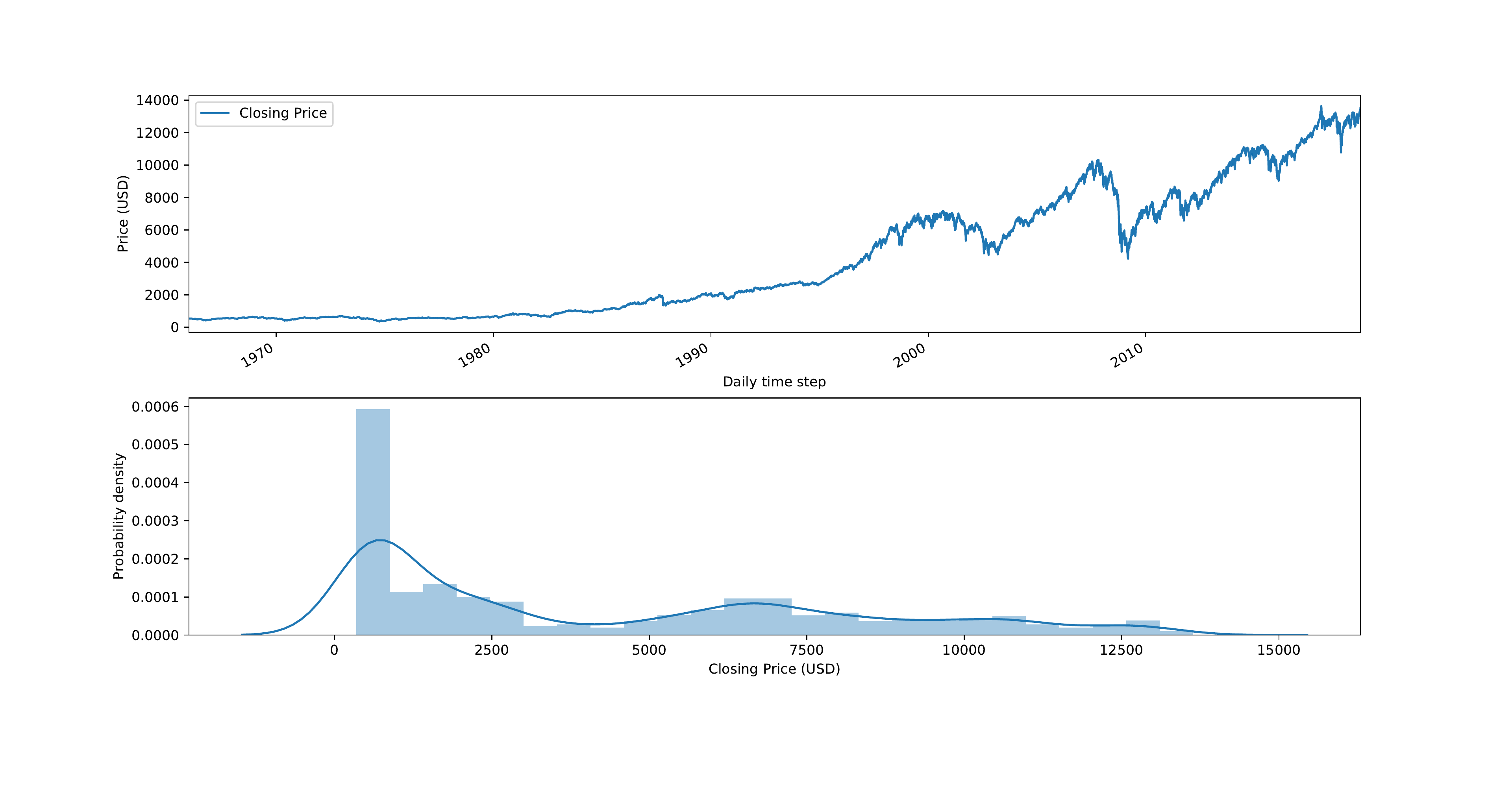}
		\caption{Top - The composite New York Stock Exchange (NYSE) closing price dataset. Bottom - Probalility distribution of the NYSE dataset.}
		\caption{Raw NYSE dataset.}
		\label{fig:data-nyse}
	\end{figure}
	
	\begin{figure}
		\centering
		\includegraphics[width=\textwidth]{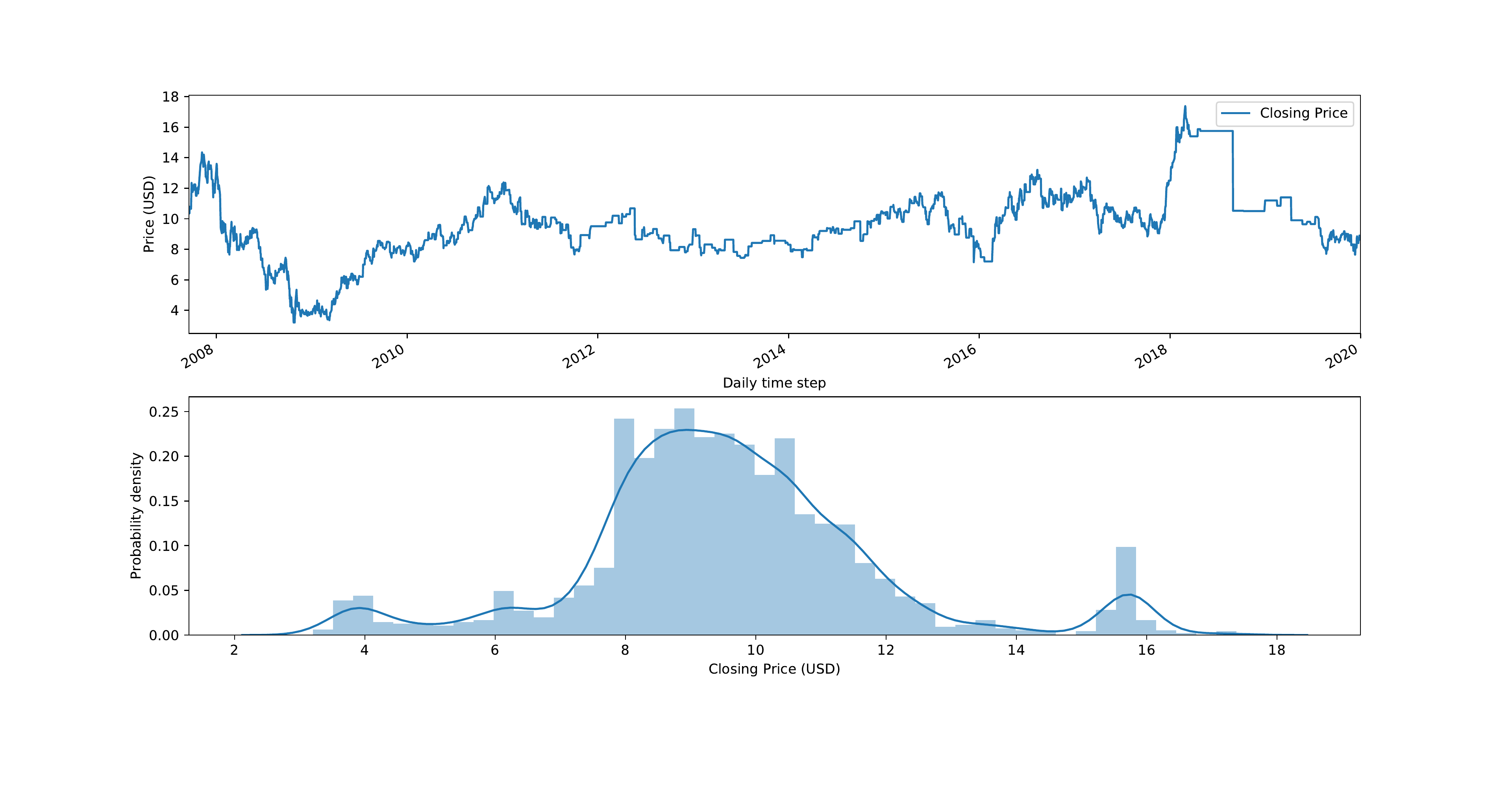}
		\caption{Top - Composite Johannesburg Stock Exchange (JSE) closing price dataset.  Bottom - Probalility distribution of the JSE dataset.}
		\label{fig:data-jse}
	\end{figure}
	
	\begin{table}[!htbp]
		\caption{Summary of the basic statistics of the segmented datasets and the input vector size per feature type.}
		\label{tab:exp-dataset-stats}
		\centering
		\resizebox{\textwidth}{!}{
			\begin{tabular}{lSSSS}
                \toprule
				& {\textbf{Voltage}} & {\textbf{Methane}} & {\textbf{NYSE}} & {\textbf{JSE}}\\
				
				\midrule
				{{\textit{Number of local raw data points}}} & {2075259} & {41786} & {13563} & {3094}\\
				
				\midrule
				{{\textit{Number of trend lines}}} & {42280} & {4419} & {10015} & {1001}\\
				
				\midrule 
				{\textit{Mean} $\pm$ {\textit{deviation of the trend slope}}} & {$-0.21 \pm 10.41$} & {$0.17 \pm 18.12$} & {$5.44 \pm 81.27$} & {$0.21 \pm 18.18$} \\
				
				\midrule
				{{\textit{Mean}} $\pm$ {\textit{deviation of the trend duration}}} & {$50.08 \pm 60.36$} & {$10.46 \pm 67.03$} & {$2.35 \pm 0.81$} & {$4.09 \pm 5.23$}\\
				
				\midrule
				{{\textit{Raw local data feature size}}} & {$19$} & {$100$} & {$4$} & {$2$}\\
				
				\midrule
				{{\textit{Raw local data + Trend line feature size}}} & {$21$} & {$102$} & {$6$} & {$4$}\\
				
				\midrule
				{{\textit{Number of data instances}}} & {$42279$} & {$4418$} & {$10014$} & {$1001$}\\
				
				\toprule	
			\end{tabular}
		}
	\end{table}
	
	\begin{table}[!htbp]
		\caption{Summary of the data instance partitioning}
		\label{tab:data-partition}
		\centering
		\begin{tabular}{lSSSS}
			\toprule
			& {\textbf{Voltage}} & {\textbf{Methane}} & {\textbf{NYSE}} & {\textbf{JSE}}\\
			\midrule
			{\textit{Number of data instances}} & {$42279$} & {$4418$} & {$10014$} & {$1001$}\\
			\midrule
			{\textit{Chosen total test sets percentage}} & {$80\%$} & {$10\%$} & {$50\%$} & {$10\%$}\\
			\midrule
			{\textit{Chosen test set size}} & {$4227$} & {$10$} & {$1001$} & {$1$}\\
			\midrule
			{\textit{Number of splits}} & {$8$} & {$44$} & {$5$} & {$101$}\\
			\midrule
			{\textit{Validation set size}} & {$4227$} & {$10$} & {$1001$} & {$1$}\\
			\midrule
			{\textit{Training set size}} & {$4227$} & {$3967$} & {$4008$} & {$899$}\\
			\bottomrule	
		\end{tabular}
	\end{table}
	
	\subsection{Model update with warm-start}\label{sec:exp-warm-start}
	The walk-forward evaluation procedure requires as many training episodes as the number of splits: one initial training and many model updates. This many training episodes can be computationally very expensive, particularly for deep neural networks. Thus, in this work, model update with warm start initialisation is used to reduce the training time of the neural network based algorithms. That is, during model update, the new network is initialised with the weights of the previous model. In effect, the patterns learnt by the previous network are transferred to the new model, therefore, reducing the number of epochs required to learn the new optimal function. In practice, the walk-forward evaluation with warm start corresponds to performing the first training with the maximum number of epochs required to converge, then using a fraction of this number for every other update. 
	This fraction - between 0.0 and 1.0 - becomes an additional hyperparameter dubbed \textit{warm start}. The lowest value that out-performed the model update without warm-start is used as the optimal value, because this technique is essentially used to speed-up the model updates. \\
	The speed-up, i.e. the expected reduction factor in the total number of epochs can be computed in advance using equation~\ref{eq:warm-start-speedup}. The equation~\ref{eq:warm-start-speedup} is derived from equation~\ref{eq:warm-start-epoch} and ~equation~\ref{eq:warm-start-epoch-}.
	\begin{equation}\label{eq:warm-start-epoch}
	E' = E + E \times (S - 1) \times \omega
	\end{equation}
	\begin{equation}\label{eq:warm-start-epoch-}
	E' = E \times ( 1 + (S - 1) \times \omega)
	\end{equation}
	\begin{equation}\label{eq:warm-start-speedup}
	speed \mbox{-} up = \frac{E}{E'} = \frac{S}{1 + (S - 1) \times \omega}
	\end{equation}
	Where, $E' \rightarrow$ \textit{Total epochs with warm start}, $E \rightarrow$ \textit{Epochs per split without warm-start}, $S \rightarrow$ \textit{Number of data partition splits}, $\omega \rightarrow $ \textit{warm-start fraction}.
	
	\begin{table}
		\caption{Our Optimal TreNet hyperparameters found by manual  experimentation. "?" means \textit{unknown} and $S \: = \{300, 600, 900, 1200\}$ }
		\label{tab:trenet-hyper}
		\centering
		\resizebox{\textwidth}{!}{
			\begin{tabular}{l c c c c c c c c c}
				
				\toprule
				& {\textbf{Dropout}}& {\textbf{L2}} & {\textbf{LR}} & {\textbf{LSTM Cells}} & {\textbf{CNN Filters}} & {\textbf{Fusion Layer}} &{\textbf{Batch Size}} & {\textbf{Epochs}} & {\textbf{Warm start}}\\
				\midrule
				\textit{Voltage} & $0.0$& 5e-4 & 1e-3 & $[600]$ & [16, 16] & 300 & $2000$ & $100$ & $0.2$\\
				\midrule
				\textit{Methane} & $0.0$& 5e-4 & 1e-3 & $[1500]$ & [4, 4] & 1200 & $2000$ & $2000$ & $0.1$\\
				\midrule
				\textit{NYSE} & $0.0$& $0.0$ & 1e-3 & $[600]$ & [128, 128] & 300 & $5000$ & $100$ & $0.5$\\
				\midrule
				\textit{JSE} & $0.0$& $0.0$ & 1e-3 & $[5]$ & [32, 32] & 10 & $500$ & $100$ & $0.05$\\
				\midrule
				\textit{Lin et al. }\cite{Lin2017} & $0.5$& 5e-4 & $?$ & $[600]$ & [32, 32] & $from \: \: S$ & $?$ & $?$ & $N/A$\\
				\bottomrule
			\end{tabular}	
		}
	\end{table}
	
\begin{table}[!htbp]
	\caption{Hyperparameters optimised for the vanilla DNN algorithms and their optimal values found for each dataset}
	\label{tab:hyperparam-vanilla-dl}
	\resizebox{\textwidth}{!}{
		\begin{tabular} {llcccc}
			\toprule
			&  & Voltage & Methane & NYSE & JSE\\
			\midrule
			\multirow{7}{*}{\textit{MLP}} 
			& {batch size} & 4000 & 250 & 5000 & 250 \\
			& {warm start}  & 0.1 & 0.1 & 0.7 & 0.05\\
			& {learning rate}  & 1e-4 & 1e-3 & 1e-3 & 1e-3\\
			& {dropout}  & 0.0 & 0.0 & 0.0 & 0.0\\
			& {weight decay}  & 0.0 & 0.0 & 5e-4 & 0.0 \\
			& {number of epochs}  & 10000 & 15000 & 500 & 100\\
			& {layer configuration} & [500, 400, 300] & [500, 400] & [500, 400, 300] & [100]\\
			\midrule
			\multirow{7}{*}{\textit{LSTM}} 
			& {batch size}  & 4000 & 2000 & 5000 & 1000 \\
			& {warm start}  & 0.1 & 0.1 & 0.01 & 0.05\\
			& {learning rate}  & 1e-2 & 1e-4 & 1e-3 & 1e-3\\
			& {dropout}  & 0.0 & 0.0 & 0.5 & 0.5 \\
			& {weight decay}  & 0.0 & 0.0 & 5e-5 & 0.0 \\
			& {number of epochs}  & 1000 & 15000 & 100 & 100\\
			& {cell configuration} & [600] & [600, 300] & [100] & [100]\\ 		 
			\midrule
			\multirow{10}{*}{\textit{CNN}}
			& {batch size}  & 2000 & 250 & 5000 & 1000 \\
			& {warm start}  & 0.5 & 0.3 & 0.4 & 0.1\\
			& {learning rate}  & 1e-3 & 1e-3 & 1e-3 & 1e-3\\
			& {dropout}  & 0.0 & 0.0 & 0.0 & 0.0 \\
			& {weight decay}  & 5e-5 & 5e-4 & 0.0 & 0.0 \\
			& {number of epochs}  & 15000 & 1000 & 12000 & 100\\
			& {filter configuration}  & [16] & [32, 32] & [32] & [32, 32]\\ 
			& {kernel configuration}  & [2] & [2, 4] & [1] & [1, 1]\\ 
			& {Pooling type}  & Max & Max & Identity & Identity\\
			& {Pooling size}  & 2 & 5 & N/A & N/A\\				 
			\bottomrule
		\end{tabular}
	}
\end{table}

\begin{table}
	\caption{Optimal hyperparameters of the traditional ML algorithms.}
	\label{tab:hyperparam-trad-ml}
	\centering
	\begin{tabular} {lccccc}
		\hline
		Algorithm & Hyperparameter & Voltage & Methane & NYSE & JSE\\
		\hline
		\multirow{4}{*}{\textit{RF}} 
		& {number of estimators} & 50 & 50 & 200 & 100 \\ 
		& {maximum depth}  & 2 & 10 & 1 & 1 \\
		& {bootstrap} & 2000 & False & True & False \\
		& {warm start} & False & False & True & True \\
		\hline
		\multirow{3}{*}{\textit{GBM}} 
		& {bootstrap type}  & gbdt & gbdt & gbdt & gbdt \\ 
		& {number of estimators} & 1 & 10000 & 1 & 4\\
		& {learning rate} & 2000 & 0.1 & 0.2 & 0.1\\	
		\hline
		\multirow{2}{*}{\textit{SVR}} 
		& {gamma} & 0.1 & 1e-4 & 1e-1 & 1e-4\\
		& {C} & 4 & 10000 & 100 & 500\\					  			  
		\hline	
	\end{tabular}
\end{table}
	
	\begin{table}[!htbp]
		\caption{Performance of vanilla DNN algorithms on raw data alone and raw data and trend line features.}
		\label{tab:feature-type-dnn}
		\begin{subtable}{1\textwidth}
			\centering
			\resizebox{\textwidth}{!}{
				\begin{tabular}{@{} l*{7}{S} @{}}
					\toprule
					& \multicolumn{3}{c}{Voltage} & \multicolumn{3}{c}{Methane}\\
					\cmidrule(lr{.75em}){2-4} \cmidrule(lr{.75em}){5-7} 
					
					& {Slope} & {Duration} & {Average} & {Slope} & {Duration} & {Average}\\
					
					\midrule
					{\textit{Raw data}}  
					& {$9.04 \pm 0.06$} & {$62.82 \pm 0.04$} & {$35.93 \pm 0.05$} 
					& {$14.57\pm 0.10$} & {$49.79 \pm 4.85$}  & {$32.18 \pm 2.47$}\\
					
					\midrule
					{\textit{Raw data + Trend lines}} 
					& {$\textbf{9.03} \pm\textbf{ 0.06}$} & {$62.81 \pm 0.04$} & {$\textbf{35.92} \pm \textbf{0.05}$} 
					& {$\textbf{14.56} \pm \textbf{0.19}$} &{$\textbf{34.46} \pm \textbf{2.79}$} & {$\textbf{24.51} \pm \textbf{1.49}$}\\
					\bottomrule	
				\end{tabular}
			}
		\end{subtable}
		
		\begin{subtable}{1\textwidth}
			\centering
			\resizebox{\textwidth}{!}{
				\begin{tabular}{@{} l*{7}{S} @{}}
					\toprule
					& \multicolumn{3}{c}{NYSE} & \multicolumn{3}{c}{JSE} \\
					\cmidrule(lr{.75em}){2-4} \cmidrule(lr{.75em}){5-7} 
					
					& {Slope} & {Duration} & {Average} & {Slope} & {Duration} & {Average}\\
					
					\midrule
					{\textit{Raw data}} 
					& {$90.76 \pm 4.43$} & {$33.08 \pm 42.08$}  & {$61.92 \pm 23.26$} 
					& {$\textbf{19.87 }\pm \textbf{0.01}$} & {$\textbf{12.51} \pm \textbf{0.09}$}  & {$\textbf{16.19} \pm \textbf{0.05}$}\\
					
					\midrule
					{\textit{Raw data + Trend lines}}
					& {$90.45 \pm 2.55$}& {$25.34 \pm 24.09$} & {$57.90 \pm 13.32$} 
					& {$21.13 \pm 0.30$} & {$12.59 \pm 0.14$} & {$16.86 \pm 0.22$}\\
					\bottomrule	
			\end{tabular}}
			\caption{MLP}
		\end{subtable}
		
		\begin{subtable}{1\textwidth}
			\centering
			\resizebox{\textwidth}{!}{
				\begin{tabular}{@{} l*{7}{S} @{}}
					\toprule
					& \multicolumn{3}{c}{Voltage} & \multicolumn{3}{c}{Methane} \\
					\cmidrule(lr{.75em}){2-4} \cmidrule(lr{.75em}){5-7} 
					
					& {Slope} & {Duration} & {Average} & {Slope} & {Duration} & {Average}\\
					
					\midrule
					{\textit{Raw data}} 
					& {$10.30 \pm 0.0$} & {$62.87 \pm 0.0$}  & {$36.59 \pm 0.0$} 
					& {$\textbf{14.21} \pm \textbf{0.19}$} & {$56.37 \pm 1.77$}  & {$35.29 \pm 0.68$}\\	
					
					\midrule
					{\textit{Raw data + Trend lines}} 
					& {$10.30 \pm 0.0$} & {$62.87 \pm 0.0$} & {$36.59 \pm 0.0$} 
					& {$14.77 \pm 0.51$} & {$\textbf{48.03} \pm \textbf{5.74}$} & {$\textbf{31.40} \pm \textbf{3.13}$}\\
					\bottomrule	
				\end{tabular}
			}
		\end{subtable}

		\begin{subtable}{1\textwidth}
			\centering
			\resizebox{\textwidth}{!}{
				\begin{tabular}{@{} l*{7}{S} @{}}
					\toprule
					& \multicolumn{3}{c}{NYSE} & \multicolumn{3}{c}{JSE} \\
					\cmidrule(lr{.75em}){2-4} \cmidrule(lr{.75em}){5-7} 
					
					& {Slope} & {Duration} & {Average} & {Slope} & {Duration} & {Average}\\
					
					\midrule
					{\textit{Raw data}} 
					& {$86.56 \pm 0.01$}& {$\textbf{0.41} \pm \textbf{0.08}$}  & {$\textbf{43.49} \pm \textbf{0.05}$}  
					& {$\textbf{19.83} \pm \textbf{0.01}$} & {$\textbf{12.68} \pm \textbf{0.01}$}  & {$\textbf{16.26} \pm \textbf{0.01}$}\\	
					
					\midrule
					{\textit{Raw data + Trend lines}} 
					& {$\textbf{86.50} \pm \textbf{0.01}$}& {$0.47\pm 0.03$}  & {$\textbf{43.49} \pm \textbf{0.02}$} 
					& {$20.16 \pm 0.03$} & {$12.74 \pm 0.02$} & {$16.45 \pm 0.03$}\\
					\bottomrule	
			\end{tabular}}
			\caption{LSTM}
		\end{subtable}
		
		\begin{subtable}{1\textwidth}
			\centering
			\resizebox{\textwidth}{!}{
				\begin{tabular}{@{} l*{7}{S} @{}}
					\toprule
					& \multicolumn{3}{c}{Voltage} & \multicolumn{3}{c}{Methane}\\
					\cmidrule(lr{.75em}){2-4} \cmidrule(lr{.75em}){5-7} 
					
					& {Slope} & {Duration} & {Average} & {Slope} & {Duration} & {Average}\\
					
					\midrule
					{\textit{Raw data}} 
					& {$\textbf{9.24} \pm \textbf{0.10}$} & {$\textbf{62.40} \pm \textbf{0.13}$}  & {$\textbf{35.82} \pm \textbf{0.12}$} 
					& {$15.07 \pm 0.35$} & {$54.79 \pm 4.55$}  & {$34.93 \pm 2.45$} \\
					
					\midrule
					{\textit{Raw data + Trend lines}} 
					& {$33.26 \pm 19.41$} & {$90.78 \pm 53.17$} & {$62.02 \pm 36.29$} 
					& {$15.14 \pm 0.28$} & {$\textbf{37.92} \pm \textbf{4.11}$} & {$\textbf{26.53} \pm \textbf{2.20}$}\\
					\bottomrule	
				\end{tabular}
			}
		\end{subtable}
		
		\begin{subtable}{1\textwidth}
			\centering
			\resizebox{\textwidth}{!}{
				\begin{tabular}{@{} l*{7}{S} @{}}
					\toprule
					& \multicolumn{3}{c}{NYSE} & \multicolumn{3}{c}{JSE} \\
					\cmidrule(lr{.75em}){2-4} \cmidrule(lr{.75em}){5-7} 
					
					& {Slope} & {Duration} & {Average} & {Slope} & {Duration} & {Average}\\
					
					\midrule
					{\textit{Raw data}} 
					& {$89.31 \pm 1.38$} & {$12.21 \pm 12.17$}  & {$50.76 \pm 6.78$} 
					& {$\textbf{19.90} \pm \textbf{0.06}$} & {$\textbf{12.48} \pm \textbf{0.21}$}  & {$\textbf{16.19} \pm \textbf{0.14}$}\\	
					
					\midrule
					{\textit{Raw data + Trend lines}} 
					& {$90.44 \pm 1.74$}& {$14.05\pm 9.52$} & {$52.25 \pm 5.63$} 
					& {$21.41 \pm 0.33$} & {$12.71 \pm 0.15$} & {$17.06 \pm 0.24$}\\
					\bottomrule	
			\end{tabular}}
			\caption{CNN}
		\end{subtable}
		\end{table}
		
	    \begin{table}[!htbp]
		\caption{Performance of traditional ML algorithms on raw data alone and raw data and trend line features.}
		\label{tab:feature-type-ensemble}
		
		\begin{subtable}{1\textwidth}
			\centering
			\resizebox{\textwidth}{!}{
				\begin{tabular}{@{} l*{7}{S} @{}}
					\toprule
					& \multicolumn{3}{c}{Voltage} & \multicolumn{3}{c}{Methane}\\
					\cmidrule(lr{.75em}){2-4} \cmidrule(lr{.75em}){5-7} 
					
					& {Slope} & {Duration} & {Average} 
					& {Slope} & {Duration} & {Average}\\
					
					\midrule
					{\textit{Local raw data}} &
					{$9.53 \pm 0.0$} & {$\textbf{63.11} \pm \textbf{0.20}$} & {$36.32 \pm 0.10$} & {$\textbf{10.09} \pm \textbf{0.01}$} & {$20.79 \pm 0.01$}  & {$\textbf{15.44} \pm \textbf{0.01}$}\\
					
					\midrule
					{\textit{Local raw data + Trend lines}} & {$\textbf{9.35} \pm \textbf{0.0}$} & {$63.19 \pm 0.29$} & {$\textbf{36.27} \pm \textbf{0.15}$} & {$11.53 \pm 0.0$} & {$\textbf{20.73} \pm \textbf{0.01}$} & {$16.13 \pm 0.01$} \\
					\bottomrule	
			\end{tabular}}
		\end{subtable}
		
		\begin{subtable}{1\textwidth}
			\centering
			\resizebox{\textwidth}{!}{
				\begin{tabular}{@{} l*{7}{S} @{}}
					\toprule
					& \multicolumn{3}{c}{NYSE} & \multicolumn{3}{c}{JSE} \\
					\cmidrule(lr{.75em}){2-4} \cmidrule(lr{.75em}){5-7} 
					
					& {Slope} & {Duration} & {Average} 
					& {Slope} & {Duration} & {Average}\\
					
					\midrule
					{\textit{Local raw data}} & 
					{$88.75 \pm 0.17$} & {$\textbf{0.29} \pm \textbf{0.0}$}  & {$44.52 \pm 0.09$} & {$\textbf{20.21} \pm \textbf{0.0}$} & {$\textbf{12.67} \pm \textbf{0.0}$}  & {$\textbf{16.44} \pm \textbf{0.0}$}\\
					
					\midrule
					{\textit{Local raw data + Trend lines}} & 
					{$86.53 \pm 0.01$}& {$0.41 \pm 0.0$} & {$\textbf{43.47} \pm \textbf{0.01}$}  & {$22.68 \pm 0.0$} & {$12.69 \pm 0.0$} & {$17.69 \pm 0.0$}\\
					\bottomrule	
			\end{tabular}}
			\caption{RF}\label{tab:feature-type-rf}
		\end{subtable}
		
		\bigskip
		\begin{subtable}{1\textwidth}
			\centering
			\resizebox{\textwidth}{!}{
				\begin{tabular}{@{} l*{7}{S} @{}}
					\toprule
					& \multicolumn{3}{c}{Voltage} & \multicolumn{3}{c}{Methane} \\
					\cmidrule(lr{.75em}){2-4} \cmidrule(lr{.75em}){5-7} 
					
					& {Slope} & {Duration} & {Average} 
					& {Slope} & {Duration} & {Average}\\
					
					\midrule
					{\textit{Local raw data}} & {$\textbf{10.0} \pm \textbf{0.0}$} & {$62.67 \pm 0.0$}  & {$36.34 \pm 0.0$} & {$13.05 \pm 0.0$} & {$75.10 \pm 0.0$}  & {$44.08 \pm 0.0$}\\
					
					\midrule
					{\textit{Local raw data + Trend lines}} & {$10.01 \pm 0.0$} & {$\textbf{62.63} \pm \textbf{0.0}$} & {$\textbf{36.32} \pm \textbf{0.0}$} & {$\textbf{12.02} \pm \textbf{0.0}$} & {$\textbf{38.34} \pm\textbf{0.0}$} & {$\textbf{25.18} \pm \textbf{0.0}$}\\
					\bottomrule	
			\end{tabular}}
		\end{subtable}
		
		\begin{subtable}{1\textwidth}
			\centering
			\resizebox{\textwidth}{!}{
				\begin{tabular}{@{} l*{7}{S} @{}}
					\toprule
					& \multicolumn{3}{c}{NYSE} & \multicolumn{3}{c}{JSE} \\
					\cmidrule(lr{.75em}){2-4} \cmidrule(lr{.75em}){5-7} 
					
					& {Slope} & {Duration} & {Average} 
					& {Slope} & {Duration} & {Average}\\
					\midrule
					{\textit{Local raw data}}  & {$86.62 \pm 0.0$}& {$0.42 \pm 0.0$}  & {$43.52 \pm 0.0$} & {$20.08 \pm 0.0$} & {$\textbf{12.62} \pm \textbf{0.0}$}  & {$16.35 \pm 0.0$}\\

					\midrule
					{\textit{Local raw data + Trend lines}} 
					& {$\textbf{86.42} \pm \textbf{0.0}$}& {$\textbf{0.41}\pm \textbf{0.0}$} 
					& {$\textbf{43.42} \pm \textbf{0.0}$} 
					& {$\textbf{19.93} \pm \textbf{0.0}$} & {$12.65 \pm 0.0$} & {$\textbf{16.29} \pm \textbf{0.0}$}\\
					\bottomrule	
			\end{tabular}}
			\caption{GBM}\label{tab:feature-type-gbm}
		\end{subtable}

		\bigskip

	\begin{subtable}{1\textwidth}
		\centering
		\resizebox{\textwidth}{!}{
			\begin{tabular}{@{} l*{7}{S} @{}}
			\toprule
			& \multicolumn{3}{c}{Voltage} & \multicolumn{3}{c}{Methane}\\
			\cmidrule(lr{.75em}){2-4} \cmidrule(lr{.75em}){5-7} 
			
			& {Slope} & {Duration} & {Average} & {Slope} & {Duration} & {Average}\\
			
			\midrule
			{\textit{Raw data}} 
			& {$\textbf{9.32} \pm \textbf{0.0}$} & {$\textbf{62.58} \pm \textbf{0.0}$} & {$\textbf{35.95} \pm \textbf{0.0}$} 
			& {$14.98 \pm 0.0$} & {$\textbf{34.39} \pm \textbf{0.0}$}  & {$\textbf{24.69} \pm \textbf{0.0}$} \\ 
			
			\midrule
			{\textit{Raw data + Trend lines}} 
			& {$9.54 \pm 0.0$} & {$62.62 \pm 0.0$} & {$36.08 \pm 0.0$}  
			& {$17.95 \pm 0.0$} & {$34.52 \pm 0.0$} & {$26.24 \pm 0.0$} \\
			\bottomrule	
		\end{tabular}}
	\end{subtable}
	
	\begin{subtable}{1\textwidth}
		\centering
		\resizebox{\textwidth}{!}{
			\begin{tabular}{@{} l*{7}{S} @{}}
					\toprule
				& \multicolumn{3}{c}{NYSE} & \multicolumn{3}{c}{JSE} \\
				\cmidrule(lr{.75em}){2-4} \cmidrule(lr{.75em}){5-7} 
				
				& {Slope} & {Duration} & {Average} & {Slope} & {Duration} & {Average}\\
				
				\midrule
				{\textit{Raw data}} 
				& {$86.55 \pm 0.0$} & {$0.42 \pm 0.0$}  & {$\textbf{43.49} \pm \textbf{0.0}$} 
				& {$\textbf{20.01} \pm \textbf{0.0}$} & {$\textbf{12.85} \pm \textbf{0.0}$}  & {$\textbf{16.43} \pm \textbf{0.0}$}\\ 
				
				\midrule
				{\textit{Raw data + Trend lines}} 
				& {$\textbf{86.54} \pm \textbf{0.0}$} & {$0.45\pm 0.0$} & {$43.50 \pm 0.0$} 
				& {$23.27 \pm 0.0$} & {$13.19 \pm 0.0$} & {$18.23 \pm 0.0$} \\
				\bottomrule	
		\end{tabular}}
		\caption{SVR}\label{tab:feature-type-svr}
	\end{subtable}

	\end{table}
	
\end{document}